\documentclass[sigconf]{acmart}
\AtBeginDocument{%
  }

\acmYear{2026}\copyrightyear{2026}
\setcopyright{cc}
\setcctype{by}
\acmConference[ACM CAIS '26]{ACM Conference on AI and Agentic Systems}{May 26--29, 2026}{San Jose, CA, USA}
\acmBooktitle{ACM Conference on AI and Agentic Systems (ACM CAIS '26), May 26--29, 2026, San Jose, CA, USA}
\acmDOI{10.1145/3786335.3813141}
\acmISBN{979-8-4007-2415-2/26/05}

\usepackage{listings}
\usepackage{graphicx}
\usepackage[linesnumbered,ruled,vlined]{algorithm2e}
\usepackage{booktabs}
\usepackage{multirow}

\begin{document}

\title{Robust Agent Compensation (RAC): Teaching AI Agents to Compensate }


\author{Srinath Perera}
\affiliation{%
  \institution{WSO2}
  \city{Santa Clara}
  \country{USA}}
\email{srinath@wso2.com}

\author{Kaviru Hapuarachchi}
\affiliation{%
  \institution{WSO2}
  \city{Santa Clara}
  \country{USA}}
\email{kaviru@wso2.com}

\author{Frank Leymann}
\affiliation{%
  \institution{University of Stuttgart}
  \city{Stuttgart}
  \country{Germany}}
\email{frank.leymann@iaas.uni-stuttgart.de}

\author{Rania Khalaf}
\affiliation{%
  \institution{WSO2}
  \city{Santa Clara}
  \country{USA}}
\email{rania@wso2.com}

\renewcommand{\shortauthors}{Perera et al.}

\begin{abstract}
We present Robust Agent Compensation (RAC), a log-based recovery paradigm (providing a safety net) implemented through an architectural extension that can be applied to most Agent frameworks to support reliable executions (avoiding unintended side effects). Users can choose to enable RAC without changing their current agent code (e.g., LangGraph agents). The proposed approach can be implemented in most existing agent frameworks via their existing extension points. We present an implementation based on LangChain, demonstrate its viability through the $\tau$²-bench and REALM-Bench, and show that when solving complex problems, RAC is 1.5-8X or more better in both latency and token economy compared to state-of-the-art LLM-based recovery approaches.

\end{abstract}




\maketitle

\section{Introduction}
\label{sec:intro}

With the advent of language models~\cite{18}, developers now have access to a wide variety of powerful AI models that have unlocked use cases previously beyond reach. Many AI applications use agents as their building blocks. Russell et al.~\cite{52} define Agents as “\textit{entities that perceive and act upon their environment}”. In modern use, agents receive inputs, analyze data, and carry out actions by calling tools or other agents, where tools are external actions available to agents. Failures in agents or the tools agents use can make agents unreliable. In a study of agents in production based on inputs from 306 participants, Pan et al.~\cite{33} highlight that “\textit{reliability is an unsolved challenge}”.  This paper focuses on a new technique towards achieving reliable agent execution. 

In order to be more specific on what we mean by this, we first present a set of definitions and concepts that provide the relevant context for the framing and approach. We call a group of agents organized as a directed graph a \textit{graph of agents}. We call supporting middleware that helps developers build, execute, and manage such a graph of agents an \textit{agent framework}. Each agent may include other agents or tools. For example, you can create a graph of agents using agent frameworks like LangGraph~\cite{20} or CrewAI~\cite{29}. A graph of agents can be as simple as a single agent or as complex as multi-agent systems defined in Junda He et al.~\cite{30}. 

AI applications, and AI agents in particular, most often have dynamic behavior where the exact execution order is determined at runtime.  Consider common ways that developers define the logic that determines how the set of activities (agents or tools to call) within such applications or agents should be executed:   

\begin{enumerate}
    \item A graph of activities
    \item Dynamically choosing which activity to execute, one at a time.  This can be done with an agent using the Reason and Act (ReAct) pattern~\cite{3}. In this pattern, the agent determines the first activity to execute, executes it, observes the world, and then repeats this reasoning and acting loop until the goal is accomplished.
    \item Planning all the activities to take next in one shot, replanning as needed. This can be done with an agent using the “Plan-and-Execute” pattern. Often, the plan is converted into a graph of agents.
    \item A combination of the above three.
\end{enumerate}

The fact that most AI applications and agents have such dynamic behavior, due to reasoning and planning such as with the ReAct or Plan-and-Execute patterns, makes using existing recovery techniques unusable, as we will see later in the Evaluation section.  We aim to address this gap.

With this context, we can now reframe this paper’s focus more succinctly as “how can I  implement a  graph of agents \textit{that achieves reliable execution}?” Agents themselves or the tools they use can fail. The code or tools used leading up to the failure may have side effects; if execution fails, those side effects may linger. We will call this "\textit{unintended agent execution side effects due to disruptions}," or "\textit{unintended execution side effects}." We call an agent execution reliable if it never leaves "\textit{unintended execution side effects}" regardless of the outcome.  

Consider the following example in Figure 1 showing the execution of activities in an agent attempting to book a trip requiring booking a flight, a hotel, and a car. In this execution, flight and hotel get booked successfully, but the car booking fails.


As an example, consider an agent attempting to book a trip requiring booking a flight, a hotel, and a car. Consider the case where the flight and hotel get booked successfully, but the car booking fails. An agent execution failure must not leave side effects. In this case, the user has been charged for the hotel and flight, but the trip booking cannot be completed. The agent needs to undo those side effects. When failures happen, any side effects may linger unless a mechanism is provided to “undo” them. One such mechanism is known in the literature as “compensation-based recovery”~\cite{44, 51}

As such, perhaps we can handle this for agents using techniques for solving that, such as by writing code to handle recovery using abstractions like transactions, distributed transactions, compensation ~\cite{44,51} or the SAGA pattern ~\cite{13} However, we observed that agent execution often involves dynamic components, such as the ReAct pattern, whose execution order is determined at runtime. The exact recovery order is important. For example, consider an agent that failed after its execution may have already produced data that other agents used. i.e., the actions of those other agents may subsequently be invalid. Detecting such dependencies is key to recovery from such disruptions. Performing recovery actions out of order may lead to unintended execution side effects.

Asking the developer to write recovery code to recover from execution-side effects in dynamic scenarios is untenable because the developer would have to anticipate all possible execution paths in advance while writing this code. 

Using ReAct agents can work for simple cases. However, with complex cases, this is an unreliable approach because, as the original ReAct paper~\cite{3} indicates, the outcome depends heavily on the quality of the model's inherent reasoning capabilities and, crucially, any few-shot examples provided in the prompt. 

Alternatively, one could adopt a planning-based approach, where an LLM-based planner builds a plan given the problem as a prompt, carries out the plan, and iteratively refines the plan and re-executes in case of failures. One example of this approach is SagaLLM~\cite{2}. 

However, as we will see in our evaluation, a planning-based solution can be costly with complex problems (both in terms of tokens and execution time) and can lead to unexpected behaviors (e.g., unnecessary compensation). Furthermore, LLM-based planning solutions carry the risk of leaving side effects due to hallucinations. 

This paper proposes an alternative solution to this problem. We propose RAC (Robust Agent Compensation), a deterministic implementation that handles compensation-based recovery independent of the actions taken, ensuring agents will not leave the system in an inconsistent state after an agentic workflow execution.

Contributions:

\begin{itemize}
    \item We propose RAC, a log-based recovery paradigm (providing a safety net) implemented through an architectural extension that can be applied to most Agent frameworks ( as described below) to support reliable executions ( avoiding unintended side effects).
    \item Identify and extend benchmarks for testing the agent for unintended execution side effects when faced with disruptions that are not mentioned in the problem description.
    \item Use the Model Context Protocol (MCP) Specification’s extension points to describe compensation pairs, creating an interoperable way for agent frameworks to discover those pairs.
    \item In evaluation, we study outcomes: token cost and latency under different task difficulty and disruption conditions, and observe that planning-based approaches can lead to costly replanning loops when faced with unknown errors, while ReAct-based agents running within RAC, which reasons at each step, can handle them with 1.5-3X or better latency and token economy.
    \item We present evidence that decoupling recovery from the ReAct Agents can enable them to solve harder problems, uncovering strong directions for the design of a robust agent execution framework \textbf{(see subsection ~\ref{sec:uplift-ract} for details)}.
    \item We provide an open-source reference implementation of RAC~\cite{46}.
\end{itemize}

The rest of the paper is organized as follows. The next section discusses related work. Sections three and four present the proposed RAC approach, architecture, and how it can be implemented across a wide range of agent frameworks. Section 5 describes the evaluation of RAC using RELAM-Bench~\cite{1} and $\tau$²-bench~\cite{8}. Section six concludes with findings and open questions. 
\section{Related Works}
\label{sec:relworks}

In real-world systems, failures are inevitable. When executions invariably fail, they may leave unwanted side effects, thereby leaking their abstractions. Although in theory, programmers can handle any side effects via thoughtful code, the resulting code is complex and error-prone. For example, Davis~\cite{41} and Gray~\cite{11} discuss a programmer keeping a “scratchpad” to track any side effects. To ease the programmer’s burden, we need higher-level abstractions. 

The most widely known abstraction is the ACID paradigm  (Härder et al.~\cite{12}). Although started with databases, systems can now handle transactions involving any kind of resource (e.g., queues, services) using global transactions based on the Two-Phase-Commit protocol (e.g., based on XA or WS-AT~\cite{42}) . 

However, when transactions are long-lasting or highly concurrent, this can lead to cascading rollbacks or severe performance slowdowns.  As an alternative, Sagas~\cite{13} proposed breaking the transaction into smaller transactions, where each such small transaction is an  ACID transaction and provides a single compensating operation that can undo its side effects. These compensations are used to reverse any side effects of subtransactions as needed (Wächter et al.~\cite{43}). This laid the groundwork of the compensation model(Colombo et al.~\cite{16}) that we will also use within the RAC. 

Leymann et al.~\cite{40} reiterated that transactions do not work well within business processing because they are long-running. Further, Helland~\cite{15} argued that transactions do not work with scalable systems. Both suggested using a compensation model instead of ACID transactions.   Note that transaction managers (as components of DBMSs) use compensation internally to implement ACID~\cite{45}. Weerawarana et al.~\cite{50} showed how compensation works with Web Services. Later, Daraghmi~\cite{17} discussed in detail how the SAGA pattern applies to microservices. 

As mentioned in the introduction, agents complicate the problem due to their dynamic nature. Consequently, there is a pressing need for clean high-level abstractions for agents that help developers handle failures and resulting side effects. 

There are many agent frameworks and platforms such as LangGraph~\cite{20}, Semantic Kernel~\cite{21}, LlamaIndex~\cite{22}, Aflow~\cite{10}, and Agentscope~\cite{9}. However, they provide limited support for managing side effects due to failures. 

Agent frameworks like LangGraph~\cite{20} used with prompts “\textit{asking to recover}” would do their best to recover and undo failed operations, inferring the required data (from the context and tool descriptions). However, they are unreliable (Shunyu et al.~\cite{3}). When handling complex scenarios that involve retries, the context becomes more complex, increasing the likelihood that the LLM will make a mistake. Therefore, we can’t depend on LLMs to reverse any failed operations. 

Let us consider approaches for recovering from failures. In the 2005 paper, Unruh et al.~\cite{3} argue that in many agent systems, it is not possible to achieve consistent recovery in a strictly distributed sense, and that instead we should focus on recovering the system to an acceptable state. It proposes achieving this through a combination of retries and compensation. When an error occurs, agents try to recover via retry and compensation, and if recovery fails, they recursively delegate to the callee. The proposed system also includes a process pair to aid in the recovery. Both our proposed system and SagaLLM~\cite{2} are built using the principles discussed in this paper. However, the Unruh et al. paper depends on agent developers to write most of the recovery logic. 

SagaLLM~\cite{2} is a framework based on the SAGA patterns that accepts a task as a prompt and creates a new agentic workflow with transactional guarantees to execute it. It includes two phases. In the first phase, SagaLLM generates an execution plan based on the user prompt (requirement) and refines it until it meets the requirements. After validating the plan with a human, the framework uses an LLM to convert it into code using an execution framework such as LangGraph. The resulting agent workflow includes compensation pairs for each operation as well as a GlobalValidationAgent for semantic validation, constraint checking, and reasoning verification. Compensations are represented as a stack in the code, and if a failure occurs, a rollback is triggered in LIFO order. 

However, SagaLLM poses several challenges to the user. 

\begin{enumerate}
    \item If the original plan runs into problems, SagaLLM will replan. As we will see in our experiments, surprises (due to disruptions or problems with the plan) can trigger an expensive retry loop.
    \item SagaLLM depends on LLM prompts for the following operations, which are susceptible to hallucination and other LLM risks.
    \begin{itemize}
        \item Generate code for executing the plan. 
        \item Find the compensation operation for each tool operation. 
        \item Find inputs that need to be passed to the compensating tool calls. 

    \end{itemize}
    \item SagaLLM does not support the recovery of agents that run static code or agents that combine static code with prompts.
\end{enumerate}

We will revisit some of those challenges in the evaluation section. 

\begin{figure}[h] 
    \centering
    \includegraphics[width=1.0\columnwidth]{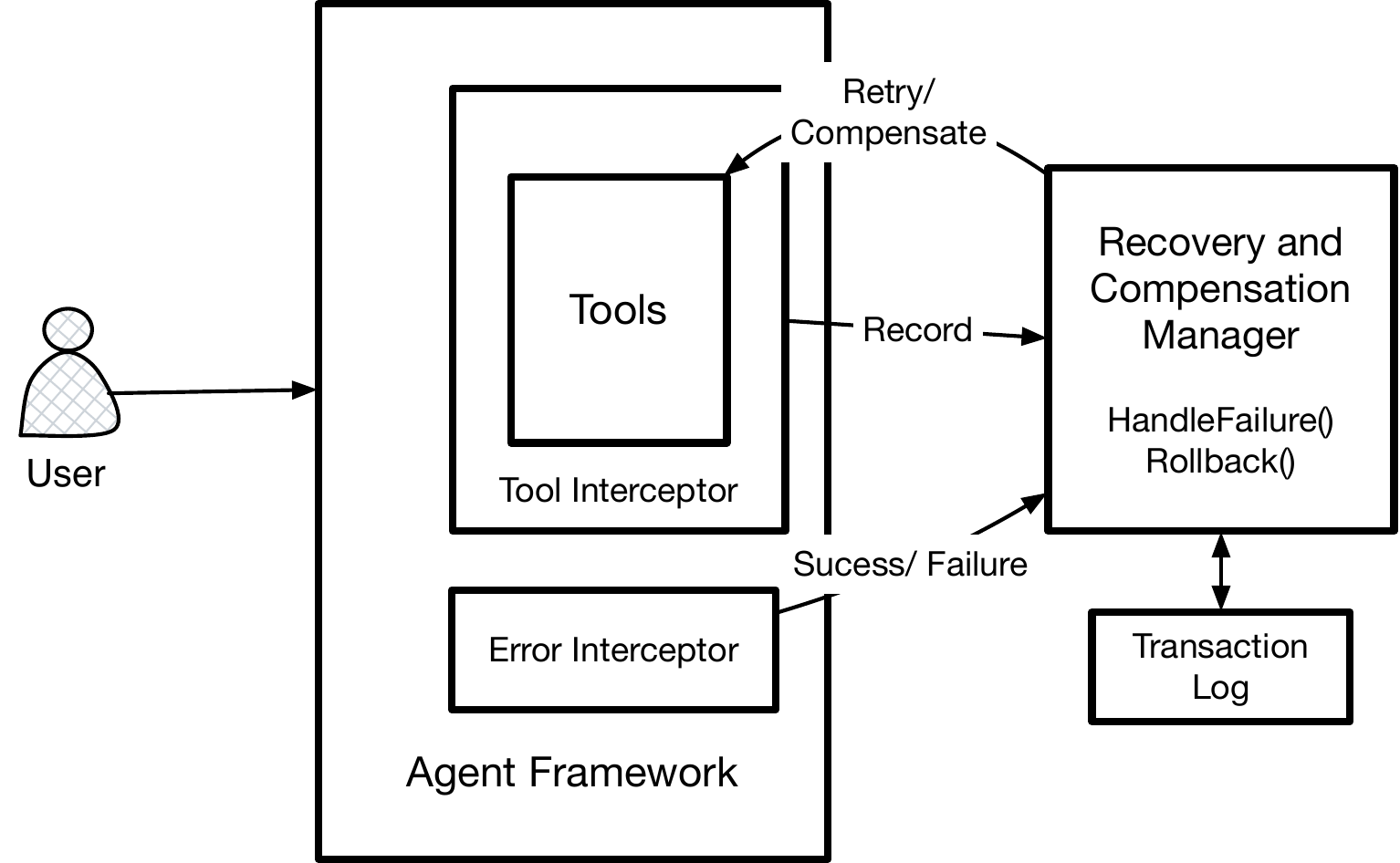} 
    \caption{RAC Architecture}
    \label{fig:rac_arch}
    \Description{RAC Architecture}
\end{figure}

To understand failure recovery, we need benchmarks. Wang et al.~\cite{7} present a benchmark “\textit{High or Hell Water}” for simulating tool failures and prompting the agent to find an alternative tool. They observe that all LLMs struggled to adapt to the errors and find a good alternative, and their performance dropped significantly. However, we could not use the benchmark to study side effects because the benchmark only includes queries (read-only operations). Similarly, the Appworld~\cite{5} benchmark is also read-only. 

$\tau$²-bench~\cite{8} is a simulation framework designed to evaluate conversational AI agents in realistic, dual-control environments where agents hand over tasks to human users as needed. We wrapped existing $\tau$²-bench tools with disruption injection, added compensation mapping, and built custom runners to benchmark compensation behaviour.

REALM-Bench~\cite{1} is a benchmark designed to test agents in real-world planning and scheduling scenarios, including environmental disruptions. We used this as our second benchmark. All REALM-Bench use cases remain solvable (can be successfully completed with retrying) in case of failures. However, a robust agent framework must gracefully handle unsolvable cases as well. Hence, we have extended the REALM-Bench benchmark to include unsolvable scenarios, which we will discuss in the evaluation section. 
\section{Robust Agent Compensation (RAC)}
\label{sec:rac}

\subsection{Proposed Design}

RAC is a recovery paradigm implemented through an architectural extension that can be applied to most Agent frameworks to support robust execution by retrying, finding alternatives, and, when recovery fails, compensating tool operations that need to be undone or compensating workflow steps running in the Agent framework. Figure~\ref{fig:rac_arch} depicts the architecture of the RAC approach. 


AI application developers enable RAC by adding interceptors to the agent framework, and then they can run their existing agents in RAC. To compensate, RAC needs to know compensation pairs for tools (another tool that can reverse this tool’s side effects; e.g., “cancelFlight” for “bookFlight”) and input mappings (how to find inputs for compensation). RAC will find them from MCP ( if tool developers have added the information to MCP tool definitions), or from definitions provided via framework API, or by asking LLM to discover them, in that order. 

The Tool Interceptor records all tool call events (e.g., start, completion, and error) in a persistent Transaction Log for each agent. Algorithm 1 shows the pseudocode for the tool interceptor. When an error occurs, the interceptor calls the \textit{handleFailure()} method of the Recovery and Compensation Manager (RCManager). Line 8 updates the agent context with a summary of rollback actions, and as discussed in subsection “How RAC uplift ReAct Agents”, updating the agent context lets follow-up agents (e.g., ReAct) solve harder problems. 

Algorithm 2 shows the pseudocode for \textit{handleFailure()}, which first retries if failures are not permanent (uses error codes and LLMs to find transient errors), then tries to find an alternative tool call (using LLM), and, in the event of failure in both, rolls back everything by calling the Rollback() method of RCManager. In line 4, RetryWithBackOff() will wait between retries to avoid overloading the servers. 
\noindent
\begin{algorithm}[tbh]
\SetKwBlock{Try}{try}{end try}
\SetKwBlock{Catch}{catch}{end catch}
\SetKwComment{Comment}{// }{}
\SetKwProg{Struct}{Struct}{}{}
\SetKwInput{KwInput}{Input}
\SetKwInput{KwOutput}{Output}
\footnotesize
\caption{Tool Interceptor}
\label{algo1}

\KwIn{tool $t$, params $p$}
\KwOut{result $r$ or handover to Recovery}
\BlankLine

\text{InvokeTool}$(t, p)$:\\
\Indp
    \Comment{1. Intercept and Record Action}
    $record \leftarrow \text{CreateRecord}(t, p, \text{status}=\text{PENDING})$\;
    \text{TransactionLog.add}{$(record)$}\;
    
        \Comment{2. Execute and Detect Errors}
        $result \leftarrow t.execute(p)$\;

        \If{$hasError(result)$}{
            $record.status \leftarrow \text{FAILED}$\;
            $report_{recovery} \leftarrow$ \text{handle\_failure}($record, error\_msg$)\;
            $updateContext(report_{recovery})$\;
            \Return{$report_{recovery}$}\;
        }
        
        $record.result \leftarrow raw\_result$\;        
        $record.status \leftarrow \text{COMPLETED}$\;
        \text{ErrorDetection.scan}{$(record)$}\;
        \Return{$raw\_result$}\;
\Indm
\end{algorithm}

For error detection, we support errors detected by the agent platform as well as semantic errors detected through user-defined prompts. Once triggered, Error Interceptor will, in turn, call the \textit{Rollback}() method of the RCManager. 

Algorithm 3 shows the pseudocode for the \textit{Rollback}() method of the RCManager. Lines 2-4 rebuild the execution graph from the Transaction Log and reverse it. Then, lines 5-13 find a compensation action for each tool call, extract the required parameters from the tool call's inputs and outputs, and invoke the compensation action.  The subsection~\ref{sec:compansation-pairs} (Design Time: Specifying Compensation Pairs) will discuss in detail how to find compensation actions and extract parameters. 

If a compensation for a tool can't be found, RAC assumes the tool has no side effects. This is a simplifying assumption that may require future work; the discussion section will revisit it. 

\medskip
\noindent
\begin{algorithm}[tbh]
\SetKwBlock{Try}{try}{end try}
\SetKwBlock{Catch}{catch}{end catch}
\SetKwComment{Comment}{// }{}
\SetKwProg{Struct}{Struct}{}{}
\SetKwInput{KwInput}{Input}
\SetKwInput{KwOutput}{Output}
\footnotesize
\caption{RCM Manager (RCM)'s Handle Failure}
\label{algo2}

\KwIn{failed\_record, error\_msg}
\KwOut{recovered result or Context for LLM}
\BlankLine

$HandleFailure(record, error)$:\\
\Indp
    \Comment{Failure context track the state}
    $Ctx_{failure} = createFailureCtx(record, error)$ \\
    \Comment{Retry for transient errors (decided via error codes)}
    \If{\textbf{not} $IsPermanentError(Ctx_{failure}$, $error)$}{
        \If{$RetryWithBackoff(record)$ \textbf{succeeds}}{
            \Return{$record.result$}\;
        }
    }
    
    \Comment{Prompt LLM to find an alternatives and run it}
    \If{{$TryAlternatives(Ctx_{failure}, record)$} \textbf{succeeds}}{
        \Return{$record.result$}\;
    }
    
    \Comment{Rollback}
    $rollback\_report \leftarrow RCM.Rollback()$\;
    
    \Comment{Return structured context for LLM Replanning}
    \Return{$FormatContextMessage(Ctx_{failure}, rollback\_report)$}\;
\Indm
\end{algorithm}

\subsection{How RAC uplifts ReAct Agents in the graph of agents}
\label{sec:uplift-ract}

As we will see in the evaluation section, the proposed RAC design has two interesting side effects when a user uses RAC to handle dynamic scenarios (e.g., by using a graph of agents that includes a ReAct Agent). First, the graph of agents (including the ReAct agent) within RAC can handle much more complex scenarios because RAC handles failures and adds the outcome of failure handling to the context. Second, RAC can handle unknown failures reliably because the ReAct agent(s) within the graph of agents reason at each step, as opposed to planning-based approaches, which get stuck in replanning loops. 



\medskip
\noindent
\begin{algorithm}[tpb]
\SetKwBlock{Try}{try}{end try}
\SetKwBlock{Catch}{catch}{end catch}
\SetKwComment{Comment}{// }{}
\SetKwProg{Struct}{Struct}{}{}
\SetKwInput{KwInput}{Input}
\SetKwInput{KwOutput}{Output}
\footnotesize
\caption{Recovery Manager(RM)'s Rollback}
\label{algo3}

\KwIn{TransactionLog (completed actions)}
\KwOut{RollbackReport}
\BlankLine

$Rollback()$:\\
\Indp
    $actions \leftarrow TransactionLog.get\_all()$\;
    $G_{execution} = buildExecuationGraph(actions)$\;

    \Comment{2. Sort activities by their order in the log to ensure that dependents are compensated before parents.}
    $plan \leftarrow TopologicalSort(G_{execution})$\;
    
    \Comment{3. Execute Compensation}
    \For{$rec$ \textbf{in} $plan$}{
        \Try{
            $c \leftarrow \text{GetCompensator}(rec.action)$\;
            $params \leftarrow ExtractParams(c, rec.result, rec.params)$\;
            $ExecuteCompensation(c, params)$\;
            $rec.status \leftarrow \text{COMPENSATED}$\;
        }
        \Catch{
            \Comment{4. If Compensation failed, we return error which will be logged and sent to the user}
            \Return{Report($error$)}\;
            
        }
    }
    \Return{Report($plan$)}\;
\Indm
\end{algorithm}

Let us consider an example. Consider a graph of agents with three agents to "edit a reservation", where the first finds the correct account using tool calls and an LLM call, the second changes the flight and hotel based on the user's inputs. and third charges a "change fee". Let's assume the last two agents are implemented as React agents. If the change to the hotel fails, RAC will first retry, then try alternatives (algorithm 2 lines 3-8), and, if all fail, undo the flight change, add everything that happened to the context, and return control to the ReAct agent. Now, the last ReAct agent will see the context of what happened and decide what to do based on the prompt (not to charge the change fee). However, because RAC has already handled the failure, ReAct does not have to handle it. Vanilla ReAct can get confused when it tries to handle complex failures and make mistakes. 

Hence, we say RAC will let ReAct agents within the graph of agents handle more complex scenarios.

\subsection{Design Time: Specifying Compensation Pairs}
\label{sec:compansation-pairs}

While running the compensation logic, the recovery manager (RCManager) needs to take a compensation action for all or a subset of each executed action.  To do that, RCManager needs to find compensations (algorithm~\ref{algo3}, line 7) and then find the inputs for the compensation operations using the inputs and outputs from earlier operations (defined as ExtractParams in pseudo code, algorithm~\ref{algo3}, line 8).  

For example, the “book flight” action can have a compensation action pair called “cancel flight,” and as the input to the cancel flight action, we need to pass the Confirmation reference from the book flight response.  

RAC looks for compensation actions and input mappings in the following order: 
\begin{enumerate}
    \item RAC will look for compensation actions and input mappings passed to the agent framework API configurations 
    \item If that is not found, RAC will check tool MCP definitions for compensation actions and input mappings ( defining those in MCP is a one-time task done by tool developers)
    \item if not found, RAC will prompt the LLM to find the compensation actions and input mappings
\end{enumerate}

If RAC finds compensation pairs but not input mappings, it will prompt the LLM to determine which outputs from previous steps map to the recovery action's inputs. Considering the example in Listing 1, RAC will prompt the LLM for the mapping ID for confirmation, providing all inputs and outputs from previous tool calls as context. Complex workflows could exhaust the LLM context size, and we can handle such cases using the RAG pattern~\cite{31}. 

Let's explore first two choices in detail. 

\textbf{Choice 1: Via Agent Framework API}

As shown in Code Listing 1, AI application  developers can define  compensation pairs and optionally their inputs as parameters to the agent creation step when using a particular agent framework. Here, lines 5-8 define compensating actions defined with LangGraph, and lines 9-14 define how to find inputs for compensating actions. 

\begin{lstlisting}[language=Python, caption=Agent Framework API Example, basicstyle=\footnotesize\ttfamily, numbers=left,label=list1]
agent = create_comp_agent(
    model="gpt-4o",
    tools=[book_flight, cancel_flight, book_hotel, 
        cancel_hotel],
    compensation_pairs={
        "book_flight": "cancel_flight",
        "book_hotel": "cancel_hotel",
    },
    state_mappers={
        "book_flight": input, 
            result: {"booking_ref": 
                result["confirmation_ref"]},
        "book_hotel": input, 
            result: {"res_id": result["reservation_id"]}
    },
)
\end{lstlisting}

\textbf{Choice 2: Via Model Context Protocol (MCP) Annotation }

It may be desired to define the compensation pair for a specific tool directly as part of the tool definition, making it available to any agent that uses it. The MCP specification \cite{37} provides an interoperable way for AI agents to dynamically discover tools. Organizations may list available tools via an MCP server, and AI Agents configured with that server may dynamically discover and use them. 

To add RAC compensation pairs to tools, we support using an annotation in the tool’s MCP schema. The following example shows an annotation that describes compensation pairs. 

\begin{lstlisting}[language=Python, caption=Example Annotation in the MCP, basicstyle=\footnotesize\ttfamily, numbers=left,label=list2]
[{
  "name": "book_flight",
  "description": "...",
  "inputSchema": {
    "type": "object",
    "properties": {
      "flight_id": { "type": "string", 
        "description": "flight number" },
      "seat_class": { "type": "string", 
        "enum": ["economy", "business", "first"] },
      "passenger_id": { "type": "string" }
    },
    "required": ["flight_id", "seat_class", "passenger_id"]
  },
  "annotations": {
   "x-compensation-tool": "cancel_flight"
  }
}]
\end{lstlisting}

\begin{lstlisting}[language=Python, caption=MCP Schema Diff, basicstyle=\footnotesize\ttfamily, numbers=left,label=list3]
[{ "op": "add", "path": "/properties/x-compensation-tool", 
    "value": { 
      "type": "string"
    } ,
    "input-mapping": { 
      "type": "string"
    } 
}]
\end{lstlisting}

Code Listing 3 shows JSON Patch instructions defined againstToolAnnotations type in the MCP Schema \cite{47}, which describes our changes. 

The above JSON  Patch adds an annotation called "x-compensation-tool. " The value of the annotation is a string that should match the name of another tool defined in the same MCP server that will act as the compensation action for the given tool.  

Similar to when compensation pairs were defined in the agent creation step, if the annotation only provides compensation tool mappings but does not provide input mappings, RAC will prompt the LLM to find the input mappings. 

This design gives a developer a choice; she can leave the agent's code unmodified using MCP-based compensation definitions or by letting LLM find compensations. Or otherwise, he can define compensations via the Agent Framework API. In either case, compensation logic executes as a post-hoc guarantee layer triggered by tool-level failure signals, timeouts, or other failure events, independent of the agent's error-handling capabilities.

\section{Implementation}
\label{sec:implementation}

\subsection{RAC Implementation}

The Recovery and Compensation Manager (RCManager) is agent-agnostic and contains the bulk of the logic. We implemented the language-agnostic part of RAC (Recovery and Compensation Manager) in Python 3, and you can find the Implementation at \cite{46}. A framework that uses RAC only needs to implement a Tool Interceptor and an Error Interceptor. Once those are implemented, RAC algorithms 1,2,3 will handle recovery. 

\subsubsection{LangGraph-based Implementation}

We implemented RAC with LangGraph, and we implemented a Tool Interceptor and an Error Interceptor using extension points in LangGraph architecture that let us intercept the pre- and post-tool invocation lifecycle and errors that occur within LangGraph.

\subsubsection{Using Extension Points in Agent Frameworks}

Similar to LangGraph, frameworks such as Semantic Kernel (Python) \cite{21}, LlamaIndex \cite{22}, Haystack \cite{23}, OpenAI Agents SDK \cite{24}, AutoGen\cite{25}, Griptape \cite{26} support hooks that let us intercept tool calls and errors, which we can use to implement a tool interceptor and an error interceptor as RAC needs. 

\subsubsection{Supporting other Agent Frameworks}

When an agent framework does not support a hook to intercept tool and error conditions, we can support RAC by wrapping each tool using a decorator. For example, a developer can extend the framework by wrapping each tool with a decorator that intercepts tool calls, tracks them, and recovers them as needed. To handle errors at the agent level, we can modify the code driving the agentic workflow to notify the recovery and compensation manager. A well-designed framework will allow adding such a decorator and intercept errors by editing a single or small part of the code base. Furthermore, in such cases, we can store the transaction log using an external database. We have assessed frameworks  PydanticAI \cite{27}, smolagents \cite{28}, and CrewAI \cite{29} for this approach and confirmed feasibility. 

Similarly, we can support non-Python frameworks by writing the decorator in a native language (e.g., Go, Java) and calling the recovery and compensation manager running as a service.

\section{Evaluation}
\label{sec:eval}

In the evaluation, as an RAC-based implementation, we ran benchmark prompts in a vanilla LangGraph ReAct Agent with RAC enabled via extension points. We evaluated  RAC against the following approaches: 
\begin{enumerate}
    \item SagaLLM - state-of-the-art solution as discussed in \cite{2}
    \item LG - Vanilla ReAct Agent (LangGraph) - ReAct pattern
    \item LG(PE) - ReAct Agent (LangGraph) with Prompt Engineering - react pattern with prompt asking to recover from failures with examples.  You can find the prompts used in \cite{46}.
\end{enumerate}

The last two are baselines often used by developers. 

For experiments, we used a machine with an M3 Pro CPU, 18GB RAM, running macOS. As LLM, we used gemini-2.5-flash unless otherwise specified. All frameworks used identical tool sets and their rollback/compensation tools. Unless otherwise specified, RAC used an LLM to identify compensation actions and input mappings (choice 3). Table~~\ref{tab:p_tasks_results} includes an additional row (RAC\_M) with ablations using manual mapping. We repeated each problem three times and collected execution time, token efficiency, and goal completion rate. We limited each problem to one million tokens. We set SagaLLM to a maximum of three planning iterations, but the discussion describes results from additional tests without those limits. 

\subsection{SagaLLM Implementation}

We encounter failures when running SagaLLM code referenced in Longling et al.\cite{47}. We wrote to the authors, but received no response. We have made the following modifications, using prompts given the papers and doing our best to follow what is described in the SagaLLM papers. The modifications are implementing phase 2 initiation per Algorithm 1 in Longling et al., improving prompts to support new LangGraph V1, and fixing generated code errors in LLM-based LangGraph generation. The modified implementation is available at \cite{46}. We also provide the changed SagaLLM code as open source via~\cite{46}, enabling interested practitioners to verify. 

\subsection{Benchmark Selection }

As discussed in the related work section, the scenarios in both Wang et al. \cite{7} and Appworld \cite{5} do not have side effects. Hence, we can’t use them for evaluating RAC. We used the following two benchmarks for evaluation.  

$\tau$²-bench \cite{8} is a benchmark for evaluating conversational AI agents in realistic customer service scenarios across three domains: Airline (e.g., handling flight cancellations, booking modifications, cabin upgrades, baggage inquiries), retail (e.g., product exchanges, order modifications, returns), and telecom (e.g, mobile data troubleshooting scenarios). The benchmark includes disruptions and allows the agent framework to abstain from some problems (as described in the prompt), and the framework may achieve success either by completing the task or abstaining without leaving side effects. Some problems have additional criteria (e.g., verify the order ID), and the frameworks must adhere to these criteria to be successful. 

The REALM-bench \cite{1} is designed to test agents in real-world planning scenarios, spanning five categories. Scheduling, routing, logistics, disaster relief, and supply chain, each incorporating domain-specific constraints, resource dependencies, and configurable disruption scenarios. It includes 11 tasks that progress from basic to highly advanced. 

Both benchmark problems increase complexity with problem numbers ( however, results suggest that problem P11 is easier). REALM-bench problems are harder than $\tau$²-bench problems and carry a significant planning component. 

As discussed in Subsection Part 2, none of these benchmarks have dynamic failures not mentioned in the prompts. We have extended both benchmarks by adding three problems with dynamic failures. 

Among them, the $\tau$²-bench provides problems and tool implementations for those problems, including fault injection. REALM-Bench only provides problem descriptions and does not provide tool implementations.  We used REALM-Bench scenarios by implementing the tools required to run them.

The following tables 1, 2, and 3 show Benchmark Results. LG represents LangGraph, and LG(PE) represents LangGraph with Prompt Engineering. Time represents the time taken by the task. a/r in the "Comp" field indicates that the problem must have minimal r compensation, but execution provided compensation. We have extensively analyzed the execution logs and, following discussion, incorporated those observations. 

\subsection{Part 1: Task with Predictable Failures}

In the first part of our implementation, we focus on “Task with Predictable Failures.” 

For this part, we use all scenarios from the $\tau$²-bench, which involve disruptions that can cause side effects. From REALM-Bench we selected scenarios 5,6,8,9, and 11 from REALM-Bench, considering Disruption Density (ability to inject failures), adaptation requirements ( how hard to recover), and state complexity. The selected scenarios are the same as those used by SagaLLM in its evaluation. 

All the above scenarios have well-defined (predictable) failures because their problem descriptions explicitly mention each failure scenario; hence, the LLM knows about them and can incorporate them into planning.

\begin{table}[htbp]
\centering
\footnotesize

\begin{tabular}{llcccc}
\toprule
\textbf{Category} & \textbf{Framework} & \textbf{Success \%} & \textbf{Compl. \%} & \textbf{Time (s)} & \textbf{Tokens} \\
\midrule
\multirow{3}{*}{Airline} & LG & 82 & - & 52 & 48k \\
                         & LG(PE) & 82 & - & 69 & 62k \\
                         & SagaLLM & 97 & - & 76 & 80k \\
                         & RAC & 97 & 81* & 50 & 77k \\
\midrule
\multirow{3}{*}{Retail}  & LG & 90 & - & 28 & 49k \\
                         & LG(PE) & 91 & - & 30 & 52k \\
                         & SagaLLM & 98 & - & 65 & 71k \\
                         & RAC & 100 & 89* & 22 & 66k \\
\midrule
\multirow{3}{*}{Telecom} & LG & 45 & 1 & 16 & 49k \\
                         & SagaLLM & 100 & 2 & 32 & 100k \\
                         & RAC & 99 & 99 & 58 & 176k \\
\bottomrule
\end{tabular}
\caption{Performance Metrics for TAU Benchmark Tasks(results from 3 Iterations)}
\label{tab:tau_metrics}
\end{table}






\subsubsection{$\tau^2$-bench Results}
Table ~\ref{tab:tau_metrics} depicts the results. $\tau$²-bench prompts include system-level instructions to abstain when there is uncertainty, which state, “\textit{you should transfer the user to a human agent if and only if the request cannot be handled within the scope of your actions.}”. Hence, a framework may be successful (represented as "Success \%") either by abstraining or by completing the task. "Compl. \%" represents the percentage of runs that were fully completed. However, $\tau$²-bench only reports this for the telecom domain. 

Success ratios increase as expected, with RAC leading or closely behind. For retail and telco, RAC is faster across the board, and it is more token-efficient than SagaLLM. RAC has good full completion rates across the board, but we do not have numbers to compare with other frameworks because $\tau^2$ bench provides those numbers only for telco. We obtained RAC numbers using internal logs (indicated via *). 

RAC completed 99\% of the telecom benchmark, while others mostly abstained, and this is likely the reason RAC used more tokens. We believe this behavior happens due to recovery prompts in the RAC. 


These results suggest that RAC is comparable to or better than other frameworks with relatively easier problems. Let us next focus on harder problems. 

\subsubsection{REALM-Bench Results}

\begin{table}[tpb]
\centering
\footnotesize
\begin{tabular}{llcrrc}
\toprule
\textbf{ID} & \textbf{Framework} & \textbf{Result} & \textbf{Tokens} & \textbf{Time (s)} & \textbf{Comp.} \\ 
\midrule
\multirow{5}{*}{P5}  & LG & 3/3 & 11k & 17 & --- \\
                     & LG(PE) & 3/3 & 28k & 24 & --- \\
                     & SagaLLM & 3/3 & 250k & 646 & --- \\
                     & RAC & 3/3 & 10k & 15 & --- \\                     
                     & RAC\_M & 3/3 & 14k & 18 & --- \\                     
\midrule
\multirow{5}{*}{P6}  & LG & 3/3 & 6k & 17 & --- \\
                     & LG(PE) & 3/3 & 6k & 14 & --- \\
                     & SagaLLM & 3/3 & 238k & 579 & 137 \\
                     & RAC & 3/3 & 9k & 15 & --- \\                     
                     & RAC\_M & 3/3 & 9k & 16 & --- \\                     
\midrule
\multirow{5}{*}{P8}  & LG & 3/3 & 6k & 17 & --- \\
                     & LG(PE) & 3/3 & 13k & 18 & --- \\
                     & SagaLLM & 3/3 & 56k & 463 & 16 \\
                     & RAC & 3/3 & 33k & 32 & 8 \\                     
                     & RAC\_M & 3/3 & 31k & 34 & 6.33 \\                     
\midrule
\multirow{5}{*}{P9}  & LG & 3/3 & 24k & 28 & --- \\
                     & LG(PE) & 3/3 & 74k & 44 & --- \\
                     & SagaLLM & 3/3 & 52k & 106 & 33 \\
                     & RAC & 3/3 & 9k & 14 & 0 \\                     
                     & RAC\_M & 3/3 & 29 & 30 & 6.00 \\                     
\midrule
\multirow{5}{*}{P11} & LG & 3/3 & 3k & 14 & --- \\
                     & LG(PE) & 3/3 & 6k & 15 & --- \\
                     & SagaLLM & 2/3 & 67k & 302 & 0 \\
                     & RAC & 3/3 & 12k & 20 & 0 \\                     
                     & RAC\_M & 3/3 & 38k & 54 & 3.33 \\                     
\bottomrule
\end{tabular}
\caption{REALM-Bench Tasks (results from 3 Iterations)}
\label{tab:p_tasks_results}
\end{table}

Table 2 shows the results for REALM-Bench. Almost all frameworks are successful except (P6 for RAC and P11 for SagaLLM). However, the cost of planning is seen in token usage and time. 

LG(PE) is about 1-3X more expensive than LG. RAC token usage is
much lower than SagaLLM and in the same range as LG and LG(PE), too, and has the best latency in most cases. SagaLLM is about 3-20 times more expensive than LG and RAC. Likely reasons could be that REALM-Bench problems are harder, and as problems become longer, we start to see the cost of pre-planning. SagaLLM ends up taking more time and tokens to solve the problems. As we have observed in traces, this happens because when problems are complex, the planning process considers many alternatives, and the plan tries to handle all cases.

SagaLLM does a lot more compensation than others. We believe the likely reasons are as follows. If the planning is successful, this stage should not lead to any replanning. However, we observed that sometimes replanning may be triggered by the Global Validator Agent due to the problems in the initial plan or the generated node graph during the code generation phase. Also, because it has to do replanning when unexpected disruptions happen, it ends up doing
a lot of unnecessary compensations due to replanning. In contrast, RAC is only obligated to handle failures that have occurred, and thus does not have to spend tokens on all potential failure cases.

\subsection{Part 2: Tasks with Dynamic Failures}

The second part of the evaluation focuses on scenarios with Dynamic Failures. We consider a scenario to have dynamic failure when the problem description does not explicitly list all possible failure scenarios (e.g., a machine temporarily breaking down or a payment being rejected). It is worth noting that, because listing all failure cases is tedious, real-world prompts often exhibit dynamic failures. 

We extended $\tau$²-bench and REALM-bench with the following new scenarios that introduce dynamic failures, extending the benchmark implementation to inject disruptions. 
\begin{enumerate}
    \item P12: Extends P11, which asks to schedule jobs on three machines. Each machine can have temporary disruptions, and this scenario tests whether the task can recover from failure by retrying. This problem can be solved by retrying.
    \item P13: Extend P12 to have permanent disruptions, and this test determines whether the framework can handle an unrecoverable scenario gracefully and compensate.
    \item P14: The "Grand Rollback" (Group Booking) - the agent needs to book unrelated flights for 3 people one by one. However, the system is rigged to fail on the 3rd booking. The agent must cancel the first 2 bookings to leave the system without side effects. This problem introduces complex multi-step compensation that requires undoing previous successful actions.
\end{enumerate}
Extended $\tau$²-bench+[48] and REALM-bench+[46] are available in~\cite{46}. 

\begin{table}[htbp]
\centering
\footnotesize
\begin{tabular}{llcrrc}
\toprule
\textbf{ID} & \textbf{Framework} & \textbf{Result} & \textbf{Tokens} & \textbf{Time (s)} & \textbf{Comp.} \\ 
\midrule
\multirow{5}{*}{P12} & LG & 0/3 & 1k & 11 & --- \\
                     & LG(PE) & 3/3 & 17k & 59 & --- \\
                     
                     & SagaLLM & 1/3 & 71k & 532 & 1 \\
                     & RAC\_M & 3/3 & 10k & 21 & --- \\
                     & RAC & 3/3 & 9k & 18 & --- \\
\midrule
\multirow{5}{*}{P13} & LG & 0/3 & 0k & 3 & --- \\
                     & LG(PE) & 0/3 & 66k & 43 & --- \\
                     & SagaLLM & 2/3 & 80k & 202 & 0 \\                     
                     & RAC\_M & 3/3 & 88k & 70 & 0.33 \\
                     & RAC & 3/3 & 76k & 58 & 0 \\
                     
\midrule
\multirow{4}{*}{P14} & LG & 0/3 & --- & 18 & 0 \\
                     & LG(PE) & 0/3 & --- & 17 & 0 \\
                     & SagaLLM & 1/3 & 95k & 173 & 2 \\                     
                     & RAC & 2/3 & 189k & 74 & 15 \\
                     
\bottomrule
\end{tabular}
\caption{Tasks with Dynamic Failures (results from 3 Iterations}
\label{tab:framework_eval}
\end{table}

Table \ref{tab:framework_eval} shows results for extended problems.  LG failed with all problems, while LG(PE) could only handle P12.  SagaLLM results are mixed, and in most failure cases, it ran out of tokens (1 million) or iteration count. In an additional test without 1 million token limits or iteration limits with P13, SagaLLM ends up taking 20X time and doing 34k unnecessary compensations. It got stuck in a replanning (10 replanning cycles) loop, abstaining only after spending 5 million tokens. Replanning occurs when errors take the system out of the initial plan, which is often the case with dynamic errors.

RAC handled all except one run of P14. RAC's performance is much better because it only needs to plan for conditions that arise at execution (rather than all possible errors when planning) and does not depend on replanning; rather, it relies on tracking what actually happens and compensating.

Comparing RAC vs. LG and LG(PE), although everything uses ReAct underneath with these problems, the ReAct agents running within RAC (see section 3.2) do not have to handle errors. Whenever an error happens, RAC will recover (retry or find an alternative)  or compensate and add what happened ( failure and compensation that happened) to the context (algorithm 1, line 8). So downstream React loops can process knowing what happened. Because the React loop never has to reason about errors, it appears as if the React loop is better than LG or LG(PE). 

\subsection{Part 3: Ablation with High Reason Model}

To understand how RAC behaves with higher reasoning, we have selected subset problems where frameworks ran into problems, and ran them three times with GPT-5.4, the most advanced model with full reasoning. We only selected problems that failed for this experiment to save costs. Table~\ref{tab:high_reason} shows the results. In the table, H means GPT-5 results, and L means Gemini-flash results. 

\begin{table}[htbp]
\centering
\footnotesize
\begin{tabular}{lcccccccc}
\toprule
\multirow{2}{*}{\textbf{Task ID}} & \multicolumn{2}{c}{\textbf{LG}} & \multicolumn{2}{c}{\textbf{LG(PE)}} & \multicolumn{2}{c}{\textbf{Saga}} & \multicolumn{2}{c}{\textbf{RAC}} \\
\cmidrule(lr){2-3} \cmidrule(lr){4-5} \cmidrule(lr){6-7} \cmidrule(lr){8-9}
& H & L & H & L & H & L & H & L \\
\midrule
P6  & 2/3 & 3/3 & 2/3 & 3/3 & 3/3 & 2/3 & 3/3 & 1/3 \\
P8  & 2/3 & 3/3 & 2/3 & 3/3 & 3/3 & 3/3 & 3/3 & 3/3 \\
P11 & 3/3 & 3/3 & 3/3 & 3/3 & 3/3 & 2/3 & 3/3 & 3/3 \\
P12 & 0/3 & 0/3 & 3/3 & 3/3 & 1/3 & 1/3 & 3/3 & 3/3 \\
P13 & 3/3 & 0/3 & 3/3 & 0/3 & 0/3 & 2/3 & 3/3 & 3/3 \\
P14 & 0/3 & 0/3 & 0/3 & 0/3 & 0/3 & 1/3 & 0/3 & 2/3 \\
\bottomrule
\end{tabular}
\caption{Ablation with High Reason Model}
\label{tab:high_reason}
\end{table}

The high-reasoning model has improved results in P6, P8 and P13 but also reduced results in P6, P8, P13 and P14. Our observations suggest that the reduction is due to hallucinations. In cases, high reasoning helped with SagaLLM, P6 took 500k+ tokens and 15m per task, P8 took close to 1M tokens and about 18m per task, and P11 took close to 1.5M tokens and about 20m per task. Firm conclusions will require more results, but the initial results above suggest that higher reasoning does not always lead to a clear advantage, and they further suggest that higher reasoning alone is not enough to reliably enable React to handle complex problems.

\section{Discussion}
\label{sec:disc}

Comparing and contrasting LG, LG(PE), SagaLLM, and RAC: LG depends on ReAct loop for recovering from failures. While sometimes it can recover from failures and compensate to avoid leaving side effects, its behavior is highly dependent on the problem, the nature of disruptions, and the prompt. For example, with the $\tau$²-bench where the prompt explicitly asks the framework to abstain if not sure and only leave a clean state, even LG does well (>80\% scenarios).  

LG(PE) added instructions on recovery and compensation to the prompt, and this improved results in several cases. (e.g., P12). 

SagaLLM depends on upfront planning. Replanning may happen due to problems with the plan, unexpected disruptions, or problems while converting the plan to LangGraph. Hence, even scenarios without disruptions can trigger replanning. 

As discussed, RAC handles recovery and compensation using a deterministic RCManager. If an error happens, RCmanager handles the error and adds what happened to the context, and the next ReAct evaluation will continue from there. Hence, ReAct agents running in RAC are shielded from errors and require doing much less work compared to the loops of LG and LG(PE). 

SagaLLM and RAC had very different execution times and token consumption behaviours.  If the plan is correct, SagaLLM’s pre-planning will be more efficient than ReAct-based planning, which needs to reason at each step. However, when problems become more complex and there is uncertainty (disruptions), SagaLLM has to replan, which becomes expensive. Multiple rounds of replanning sometimes lead to a lot of unnecessary compensation. We see this happening with P6, P8, and P9. Furthermore, when not all errors are mentioned in the prompts, these problems become even worse. This shows a critical limitation of planning-based approaches. Uncertainty can trigger costly replanning loops, and errors not specified in the protocol can heighten the risks. However, that does not mean planning is always bad. We believe that exploring when to use which approach and how to combine both planning and dynamic execution offer interesting opportunities for future research. 

In contrast, RAC chooses to decouple agent execution side effects from LLM reasoning, where the RC manager tracks all activities via a translation log and ensures that all side effects are compensated. On one hand, this provides the predictability and stability we expect from a robust system. Furthermore, it frees the reasoning loop from having to worry about errors, which simplifies its reasoning and increases its effectiveness. 

SagaLLM does not require users to define compensation pairs, but it depends on LLM in finding them. RAC can do the same, but optionally support providing compensation pairs via MCP or agent framework API, which will increase its robustness. Furthermore, if MCP is used, the tool developers only have to define it once, and the application developers do not have to do anything. 

If a compensation for a tool can't be found, RAC assumes the tool has no side effects. Given a tool that has side effects, but does not have a compensation action, neither RAC nor other compensation solutions (e.g., workflows) can add anything for those use cases. In this case, if RAC throws an error when it can't find a tool, this interferes with normal use. For example, consider the print(...) operation. If RAC throws an error because LLM cannot find an unprint (..) operation, it will interfere with many normal permissible scenarios. We have left exploring other solutions, e.g., passing the decision to the end user (via configuration) or using an LLM to determine whether the action has side effects, as future work.

Current RAC implementation rolls back everything. As discussed on Algorithm 3, in response to an error, RAC will retry and also try alternatives. Hence, the chance that partial compensation and retrying will succeed is small. LLM planning (e.g., ReAct, SAGALLM) might rethink and use an alternative strategy to recover, but aligning the scope-based recovery with the alternative strategy without changing current code (one of our key goals) is complicated. Hence, we have not explored adding scope to the compensation, but acknowledge that this is a good future research area.

\section{Conclusion}
\label{sec:conc}
When implementing AI applications that involve a set of actions, we need to handle failures and ensure they do not cause lingering side effects. Even though there are higher-level abstractions such as ACID and SAGA, they are challenging to use with dynamic agents like the React agent. Asking AI developers to handle failure scenarios through first principles is complex and error-prone. On the other hand, LLM-based planning approaches have been shown to be expensive and susceptible to hallucinations. 

To solve the agent execution side effects problem, we present Robust Agent Compensation (RAC), a recovery paradigm implemented through an architectural extension.  

RAC's core contribution is a post-hoc guarantee layer that can be integrated into existing agent frameworks (like LangGraph or CrewAI) via their existing extension points. Unlike previous systems that rely on the LLM to "reason" its way out of a failure, RAC records key events in a transaction log and uses them to compensate for any unintended side effects by rebuilding execution history from the log to perform a precise LIFO (Last-In-First-Out) rollback when a task becomes unrecoverable. Evaluation based on $\tau$²-bench and the RELAM-Bench shows that, for complex problems, RAC has about 1.5-8X better token Efficiency and lower latency. Furthermore, by reducing the dependency on LLMs, RAC reduces the potential for hallucination. 

Furthermore, we standardized compensation strategies by introducing syntax and semantics  (compensation pairs in Agent API and through MCP) to define how an action should be "undone," ensuring interoperability across different systems. 

Moreover, we extended the benchmarks to include "Dynamic Failures," where failures aren't explicitly described in the prompt (e.g., a random machine breakdown), and  "Unsolvable Scenarios," where an agent must undo multiple successful prior steps because a late-stage step failed and no alternatives exist. Both of these occur in real-world use cases. 

Finally, we argue that decoupling reliability from LLM reasoning, as done by RAC, provides the predictability and stability we expect from a robust system while freeing the reasoning loop from worrying about errors, improving its effectiveness, and enabling it to solve harder problems. We believe RAC unlocks a promising direction for robust agent execution design.

\bibliographystyle{ACM-Reference-Format}
\bibliography{base}

@String{Computing = "Computing" }

@String{Computer = "{IEEE} Computer" }

@String{Springer = "Springer-Verlag" }

@misc{1,
      title={REALM-Bench: A Benchmark for Evaluating Multi-Agent Systems on Real-world, Dynamic Planning and Scheduling Tasks}, 
      author={Longling Geng and Edward Y. Chang},
      year={2025},
      eprint={2502.18836},
      archivePrefix={arXiv},
      primaryClass={cs.AI},
      url={https://arxiv.org/abs/2502.18836}, 
}

@article{2,
  title={SagaLLM: Context Management, Validation, and Transaction Guarantees for Multi-Agent LLM Planning},
  author={Chang, Edward Y and Geng, Longling},
  journal={Proceedings of the VLDB Endowment},
  volume={18},
  number={12},
  pages={4874--4886},
  year={2025},
  publisher={VLDB Endowment}
}

@misc{3,
      title={ReAct: Synergizing Reasoning and Acting in Language Models}, 
      author={Shunyu Yao and Jeffrey Zhao and Dian Yu and Nan Du and Izhak Shafran and Karthik Narasimhan and Yuan Cao},
      year={2023},
      eprint={2210.03629},
      archivePrefix={arXiv},
      primaryClass={cs.CL},
      url={https://arxiv.org/abs/2210.03629}, 
}

@article{4,
  title={Measuring agents in production},
  author={Pan, Melissa Z and Arabzadeh, Negar and Cogo, Riccardo and Zhu, Yuxuan and Xiong, Alexander and Agrawal, Lakshya A and Mao, Huanzhi and Shen, Emma and Pallerla, Sid and Patel, Liana and others},
  journal={arXiv preprint arXiv:2512.04123},
  year={2025}
}

@inproceedings{5,
  title={Appworld: A controllable world of apps and people for benchmarking interactive coding agents},
  author={Trivedi, Harsh and Khot, Tushar and Hartmann, Mareike and Manku, Ruskin and Dong, Vinty and Li, Edward and Gupta, Shashank and Sabharwal, Ashish and Balasubramanian, Niranjan},
  booktitle={Proceedings of the 62nd Annual Meeting of the Association for Computational Linguistics (Volume 1: Long Papers)},
  pages={16022--16076},
  year={2024},
  publisher = {Association for Computational Linguistics},
  address   = {Bangkok, Thailand}
}

@inproceedings{6,
  title={Semantic-compensation-based recovery in multi-agent systems},
  author={Unruh, Amy and Harjadi, Henry and Bailey, James and Ramamohanarao, Kotagiri},
  booktitle={IEEE 2nd Symposium on Multi-Agent Security and Survivability, 2005.},
  pages={85--94},
  year={2005},
  organization={IEEE}
}

@misc{7,
      title={Hell or High Water: Evaluating Agentic Recovery from External Failures}, 
      author={Andrew Wang and Sophia Hager and Adi Asija and Daniel Khashabi and Nicholas Andrews},
      year={2025},
      eprint={2508.11027},
      archivePrefix={arXiv},
      primaryClass={cs.CL},
      url={https://arxiv.org/abs/2508.11027}, 
}

@misc{8,
      title={$\tau^2$-Bench: Evaluating Conversational Agents in a Dual-Control Environment}, 
      author={Victor Barres and Honghua Dong and Soham Ray and Xujie Si and Karthik Narasimhan},
      year={2025},
      eprint={2506.07982},
      archivePrefix={arXiv},
      primaryClass={cs.AI},
      url={https://arxiv.org/abs/2506.07982}, 
}

@misc{9,
      title={AgentScope: A Flexible yet Robust Multi-Agent Platform}, 
      author={Dawei Gao and Zitao Li and Xuchen Pan and Weirui Kuang and Zhijian Ma and Bingchen Qian and Fei Wei and Wenhao Zhang and Yuexiang Xie and Daoyuan Chen and Liuyi Yao and Hongyi Peng and Zeyu Zhang and Lin Zhu and Chen Cheng and Hongzhu Shi and Yaliang Li and Bolin Ding and Jingren Zhou},
      year={2024},
      eprint={2402.14034},
      archivePrefix={arXiv},
      primaryClass={cs.MA},
      url={https://arxiv.org/abs/2402.14034}, 
}

@misc{10,
      title={AFlow: Automating Agentic Workflow Generation}, 
      author={Jiayi Zhang and Jinyu Xiang and Zhaoyang Yu and Fengwei Teng and Xionghui Chen and Jiaqi Chen and Mingchen Zhuge and Xin Cheng and Sirui Hong and Jinlin Wang and Bingnan Zheng and Bang Liu and Yuyu Luo and Chenglin Wu},
      year={2025},
      eprint={2410.10762},
      archivePrefix={arXiv},
      primaryClass={cs.AI},
      url={https://arxiv.org/abs/2410.10762}, 
}

@inproceedings{11,
author = {Gray, Jim},
title = {The transaction concept: virtues and limitations (invited paper)},
year = {1981},
publisher = {VLDB Endowment},
booktitle = {Proceedings of the Seventh International Conference on Very Large Data Bases - Volume 7},
pages = {144–154},
numpages = {11},
address = {Cannes, France},
series = {VLDB '81}
}

@article{12,
  title={Principles of transaction-oriented database recovery},
  author={Haerder, Theo and Reuter, Andreas},
  journal={ACM computing surveys (CSUR)},
  volume={15},
  number={4},
  pages={287--317},
  year={1983},
  publisher={ACM New York, NY, USA}
}

@article{13,
  title={Sagas},
  author={Garcia-Molina, Hector and Salem, Kenneth},
  journal={ACM Sigmod Record},
  volume={16},
  number={3},
  pages={249--259},
  year={1987},
  publisher={ACM New York, NY, USA}
}

@article{15,
  title={Life beyond distributed transactions: an apostate’s opinion},
  author={Helland, Pat},
  journal={Queue},
  volume={14},
  number={5},
  pages={69--98},
  year={2016},
  publisher={ACM New York, NY, USA}
}

@article{16,
  title={Recovery within long-running transactions},
  author={Colombo, Christian and Pace, Gordon J},
  journal={ACM Computing Surveys (CSUR)},
  volume={45},
  number={3},
  pages={1--35},
  year={2013},
  publisher={ACM New York, NY, USA}
}

@article{17,
  title={Enhancing saga pattern for distributed transactions within a microservices architecture},
  author={Daraghmi, Eman and Zhang, Cheng-Pu and Yuan, Shyan-Ming},
  journal={Applied Sciences},
  volume={12},
  number={12},
  pages={6242},
  year={2022},
  publisher={MDPI}
}

@article{18,
  title={A survey on evaluation of large language models},
  author={Chang, Yupeng and Wang, Xu and Wang, Jindong and Wu, Yuan and Yang, Linyi and Zhu, Kaijie and Chen, Hao and Yi, Xiaoyuan and Wang, Cunxiang and Wang, Yidong and others},
  journal={ACM transactions on intelligent systems and technology},
  volume={15},
  number={3},
  pages={1--45},
  year={2024},
  publisher={ACM New York, NY}
}

@software{20,
  author = {LangChain AI},
  title = {LangGraph: Building stateful, multi-agent applications with LLMs},
  year = {2024},
  url = {https://github.com/langchain-ai/langgraph},
}

@software{21,
  author = {{Microsoft Semantic Kernel Team}},
  title = {Semantic Kernel: Integrate LLMs into your applications},
  year = {2024},
  url = {https://github.com/microsoft/semantic-kernel},
}

@software{22,
  author = {{LlamaIndex Team}},
  title = {LlamaIndex: Data framework for LLM applications},
  year = {2024},
  url = {https://github.com/run-llama/llama_index},
}

@software{23,
  author = {{deepset GmbH}},
  title = {Haystack: The open source NLP framework for composable AI},
  year = {2024},
  url = {https://github.com/deepset-ai/haystack},
}

@software{24,
  author = {{OpenAI}},
  title = {OpenAI Agents SDK},
  year = {2024},
  url = {https://github.com/openai/openai-agents-python},
}

@misc{25,
      title={AutoGen: Enabling Next-Gen LLM Applications via Multi-Agent Conversation}, 
      author={Qingyun Wu and Gagan Bansal and Jieyu Zhang and Yiran Wu and Beibin Li and Erkang Zhu and Li Jiang and Xiaoyun Zhang and Shaokun Zhang and Jiale Liu and Ahmed Hassan Awadallah and Ryen W White and Doug Burger and Chi Wang},
      year={2023},
      eprint={2308.08155},
      archivePrefix={arXiv},
      primaryClass={cs.AI},
      url={https://arxiv.org/abs/2308.08155}, 
}

@software{26,
  author = {{Griptape Team}},
  title = {Griptape: Python framework for AI workflows and pipelines},
  year = {2024},
  url = {https://github.com/griptape-ai/griptape},
}

@software{27,
  author = {{Pydantic Team}},
  title = {PydanticAI: Agent Framework for Production-Grade Generative AI},
  year = {2024},
  url = {https://github.com/pydantic/pydantic-ai},
}

@software{28,
  author = {{Hugging Face Team}},
  title = {smolagents: A tiny library to build agents that write python code},
  year = {2024},
  url = {https://github.com/huggingface/smolagents},
}

@software{29,
  author = {Moura, João},
  title = {CrewAI: Orchestrating Role-Playing, Autonomous AI Agents},
  year = {2024},
  url = {https://github.com/crewAIInc/crewAI},
}

@article{30,
  title={Llm-based multi-agent systems for software engineering: Literature review, vision, and the road ahead},
  author={He, Junda and Treude, Christoph and Lo, David},
  journal={ACM Transactions on Software Engineering and Methodology},
  volume={34},
  number={5},
  pages={1--30},
  year={2025},
  publisher={ACM New York, NY}
}

@article{31,
  title={A Survey on RAG with LLMs},
  author={Arslan, Muhammad and Ghanem, Hussam and Munawar, Saba and Cruz, Christophe},
  journal={Procedia computer science},
  volume={246},
  pages={3781--3790},
  year={2024},
  publisher={Elsevier}
}

@misc{33,
      title={Measuring Agents in Production}, 
      author={Melissa Z. Pan and Negar Arabzadeh and Riccardo Cogo and Yuxuan Zhu and Alexander Xiong and Lakshya A Agrawal and Huanzhi Mao and Emma Shen and Sid Pallerla and Liana Patel and Shu Liu and Tianneng Shi and Xiaoyuan Liu and Jared Quincy Davis and Emmanuele Lacavalla and Alessandro Basile and Shuyi Yang and Paul Castro and Daniel Kang and Joseph E. Gonzalez and Koushik Sen and Dawn Song and Ion Stoica and Matei Zaharia and Marquita Ellis},
      year={2026},
      eprint={2512.04123},
      archivePrefix={arXiv},
      primaryClass={cs.CY},
      url={https://arxiv.org/abs/2512.04123}, 
}

@online{37,
  author = {{LangChain AI}},
  title = {Model Context Protocol (MCP) Documentation},
  year = {2025},
  url = {https://docs.langchain.com/oss/python/langchain/mcp},
  note = {Accessed: 12 January 2026},
}

@book{40,
  author    = {Frank Leymann and Dieter Roller},
  title     = {Production Workflow-Concepts and Techniques},
  year      = {1999},
  publisher = {Prentice Hall},
  address   = {Upper Saddle River, NJ, USA},
  isbn      = {978-0130217530}
}

@article{41,
  title={Data processing spheres of control},
  author={Davies, Charles T},
  journal={IBM Systems Journal},
  volume={17},
  number={2},
  pages={179--198},
  year={1978},
  publisher={IBM}
}

@techreport{42,
  author      = {{OASIS WS-TX Technical Committee}},
  title       = {Web Services Atomic Transaction ({WS-AtomicTransaction}) Version 1.1},
  institution = {OASIS},
  month       = dec,
  year        = {2006},
  type        = {OASIS Standard},
  url         = {https://docs.oasis-open.org/ws-tx/wstx-wsat-1.1-spec-cd-01.pdf}
}

@book{43,
  title={Database Transaction Models for Advanced Applications},
  author={Elmagarmid, Ahmed K.},
  year={1992},
  publisher={Morgan Kaufmann Publishers Inc.},
  address={San Francisco, CA, USA},
  isbn={978-1558602144}
}

@incollection{44,
  title={Supporting business transactions via partial backward recovery in workflow management systems},
  author={Leymann, Frank},
  booktitle={Datenbanksysteme in B{\"u}ro, Technik und Wissenschaft: GI-Fachtagung, Dresden, 22.--24. M{\"a}rz 1995},
  pages={51--70},
  year={1995},
  publisher={Springer},
  address={Dresden, Germany}
}

@book{45,
  author = {Gray, Jim and Reuter, Andreas},
  title = {Transaction Processing: Concepts and Techniques},
  year = {1992},
  publisher = {Morgan Kaufmann Publishers Inc.},
  address = {San Francisco, CA, USA},
  isbn = {978-1558601901}
}

@misc{46,
  author       = {WSO2},
  title        = {Source Code and Data for RAC},
  year         = {2026},
  publisher    = {WSO2},
  howpublished = {https://github.com/wso2-incubator/research-rac}
}

@misc{47,
  author       = {Longling Geng and Edward Y. Chang},
  title        = {SagaLLM: Context Management, Validation, and Transaction Guarantees for Multi-Agent LLM Planning},
  year         = {2025},
  publisher    = {GitHub},
  journal      = {GitHub repository},
  howpublished = {\url{https://github.com/genglongling/SagaLLM}}
}

@book{50,
  author    = {Weerawarana, Sanjiva and Curbera, Francisco and Leymann, Frank and Storey, Tony and Ferguson, Donald F},
  title     = {Web services platform architecture: SOAP, WSDL, WS-policy, WS-addressing, WS-BPEL, WS-reliable messaging and more},
  year      = {2005},
  publisher = {Prentice Hall},
  address   = {Upper Saddle River, NJ, USA},
  isbn      = {978-0131488748}
}

@inproceedings{51,
  title={Revisiting the behavior of fault and compensation handlers in WS-BPEL},
  author={Khalaf, Rania and Roller, Dieter and Leymann, Frank},
  booktitle={OTM Confederated International Conferences" On the Move to Meaningful Internet Systems"},
  pages={286--303},
  year={2009},
  publisher={Springer},
  address={Rhodes, Greece}
}

@book{52,
  author    = {by Peter Norvig (Author), Stuart Russell},
  title     = {Artificial Intelligence: A Modern Approach, 4th Edition},
  year      = {2021},
  publisher = {Pearson},
  address   = {Hoboken, NJ, USA},
  isbn      = {978-1292401133}
}

\end{document}